\documentclass[final,2p,times,twocolumn,authoryear]{elsarticle}

\usepackage{amsmath}
\usepackage{amssymb}
\usepackage{caption}
\usepackage{color}
\usepackage{subcaption}
\usepackage{hyperref}
\usepackage[super]{nth}
\usepackage[all]{hypcap}
\usepackage{multirow}
\usepackage{lineno}

\newcommand{\expnumber}[2]{{#1}\times10^{#2}}

\newcommand{\eg}{\emph{e.g.},\ }
\newcommand{\ie}{\emph{i.e.},\ }

\newcommand{\etc}{\emph{etc.}\ }
\graphicspath{{./beam_bending_figures/}{./poisson_figures/}}
\journal{Computers and Structures}
\makeatletter
\def\ps@pprintTitle{%
  \let\@oddhead\@empty
  \let\@evenhead\@empty
  \let\@oddfoot\@empty
  \let\@evenfoot\@oddfoot
}
\makeatother

\begin{document}
\begin{frontmatter}



\title{Genetic Programming Based Symbolic Regression for Analytical Solutions to
Differential Equations}
\author[1]{Hongsup Oh}
\ead{hongsup.oh@utah.edu}
\author[2,3]{Roman Amici}
\ead{amicir@gmail.com}
\author[4]{Geoffrey Bomarito}
\ead{geoffrey.f.bomarito@nasa.gov}
\author[3]{Shandian Zhe}
\ead{zhe@cs.utah.edu}
\author[2,3]{Robert M. Kirby}
\ead{kirby@cs.utah.edu}
\author[1]{Jacob Hochhalter}
\ead{jacob.hochhalter@utah.edu}
\address[1]{Department of Mechanical Engineering, University of Utah}
\address[2]{Scientific Computing and Imaging Institute, University of Utah}
\address[3]{School of Computing, University of Utah}
\address[4]{NASA Langley Research Center}

\begin{abstract}
In this paper, we present a machine learning method for the discovery of
analytic solutions to differential equations. The method utilizes an inherently
interpretable algorithm, genetic programming based symbolic regression.   Unlike conventional accuracy measures in
machine learning we demonstrate the ability to recover true analytic
solutions, as opposed to a numerical approximation. The method is verified by
assessing its ability to recover known analytic solutions for two separate
differential equations. The developed method is compared to a conventional,
purely data-driven genetic programming based symbolic regression algorithm.
The reliability of successful evolution of the true
solution, or an algebraic equivalent, is demonstrated.

\end{abstract}


\end{frontmatter}

\newpage

\section{Introduction}
\label{intro}
The governing physics in engineering mechanics problems is often formalized
mathematically via ordinary or partial differential equations (ODEs or PDEs,
respectively).  However, direct analytical solutions of these equations are
generally not attainable in practice other than in idealized cases. For
practical engineering problems involving challenges such as complex geometries,
discretization-based methods, \eg finite element (FE) analysis, are utilized to
provide numerical approximations of the solution.  In recent years, data-driven
machine learning (ML) alternatives have become widespread, and methods have been
developed to incorporate knowledge of domain physics: often dubbed
theory-guided, physics-informed, or physics-regularized ML
\cite{karniadakis2021, raissi2019physics, karpatne2017theory}.  Inspired by the
work of \citet{raissi2017physics}, wherein physics-informed neural networks
(PINNs) were demonstrated for finding approximate solutions of PDEs, we
demonstrate physics-regularization within an inherently interpretable ML
method for determination of symbolic solutions to ODEs and PDEs. While the
developed method seeks minimization of a residual for approximate
symbolic solutions it is demonstrated that true analytical solutions can be
reliably discovered.

In practice, data in engineering programs are typically expensive to acquire,
especially at a scale sufficient to train most conventional ML models. This
often forces decisions based on relatively small datasets. However, recent
advances in high-throughput data acquisition methods to support ML are promising
\cite{Heckman_2020}.  Complementing data acquisition in the ML process are
physics-regularized approaches which can promote generalizable models that are
consistent with \emph{a priori} knowledge \cite{raissi2017physics,
raissi2019physics, warner2020inverse}.  Although the area of PINNs is especially
promising, black-box ML methods applied to scientific and engineering
applications face challenges related to the interpretability and explainability
of the models.  Unfortunately, opening the metaphorical ML black box does not
provide immediate insight into how or why a model works: a necessary step in
engineering and science scenarios. To remedy this opacity, the current trend is
to utilize additional tools (\eg visualization, weight-matrix analysis, or decision
trees) to aid interpretation of the resulting complex solutions
\cite{adadi2018peeking,dovsilovic2018explainable}. The knowledge-extraction
process is then analogous to current discretization-based paradigms (\eg FE):
evaluate the approximate solution at a set of points and generate a contour plot
or data table. In the context of engineering mechanics, lacking interpretability
and explainability can perpetuate an inherent lack of understanding about the
generated model and its prediction capabilities and limitations.  Furthermore,
in many cases, there is litte or no added accuracy benefit to using such
black-box models compared to interpretable alternatives \cite{rudin2019we}.  

Herein, genetic programming based symbolic regression (GPSR) is
used due to its inherent interpretability and recent successful application to
physics and mechanics problems
\cite{versino2017data,bomarito2021development,hernandez2019fast}. The standard
form of GPSR has been shown to be capable of developing true, verifiable models
in the area of solid mechanics \cite{bomarito2021development} and of discovering
underlying physics from data \cite{schmidt2009distilling}. Extending beyond
previous works, we augment standard GPSR with a physics-regularized fitness
function, PR-GPSR, to evolve solutions to known differential equations \ie where
the governing mechanics are known and a solution is sought.  GPSR produces
models in the form of analytic equations \cite{koza1992genetic}; thus, when
regularized by the residual of the known ODE or PDE, it aims to produce analytic
solutions thereof. As will be demonstrated herein for symbolic models, as is
also true of PINNs, physics-regularized ML requires only (at a minimum) a
complete statement of the differential equation and boundary or initial
conditions necessary for a well-posed problem. 

Due to the free-form symbolic regression nature of GPSR, the application of
physics-regularization becomes analogous to the determination of closed-form
solutions to differential equations and as such distinguished from more
conventional numerical approximations. For example, in a conventional
mathematical approach the practitioner proposes an ansatz space from which
possible analytic solutions can be built, as well as the allowable ways in
which expressions can be assembled (\eg addition, multiplication, composition,
\mbox{\etc)}.  In the language of ML, the building blocks or dictionaries of
mathematical functions are assembled and evolved to evaluate possible
analytical solutions.  GPSR with a physics-regularized fitness function
essentially automates this process of practitioner proposition of analytic
solutions.  In the context of engineering mechanics, a beneficial output of
this approach is that symbolic models fit naturally within existing workflows.
When coupled with user interpretation, and explainability via satisfaction of
known differential equations, this promotes increased trust, accessibility, and
transfer of generated ML models into practice.  Further, more insights readily
gained from symbolic equations can suggest future data acquisition or new
theories by identifying model characteristics \eg asymptotes.  Lastly, symbolic
equations allow for a broader mathematical treatment of produced models \eg
formulation of analytic adjoints.  

The paper is organized as follows.  In Section \ref{sec:methods}, we present a
brief overview of GPSR and provide details on how we apply GPSR for learning the
solutions to differential equations. In Section \ref{sec:experiments} we define two numerical
experiments based on boundary-value problems.  In Section \ref{sec:results}, we
demonstrate the performance of the developed method. In Section
\ref{sec:discussion}, we present a discussion of results with a summary of the
successes and challenges of the developed methodology. We conclude in Section
\ref{sec:conclusions} with a summary and a vision for next steps.

\section{Methods}
\label{sec:methods}

\subsection{Genetic Programming for Symbolic Regression (GPSR)}
\label{sec:gpsr}
Symbolic regression is a method that aims to model an input data set without
assuming its form.  Instead, candidate models are proposed and evaluated by the
algorithm, with the only assumption being that the data can be modeled by some
algebraic expression.  This approach is in contrast to traditional regression methods in
which model form selection is made first, and the regression method then
estimates the model parameters.  SR reformulates the traditional regression
problem into that of searching for an optimal model form and its associated
parameters.  Note that this is similar to what is done in deep neural networks
(DNNs), in which through a non-linear composition of layers a regressor learns
both coefficients and basis functions (model form) from data.  However, in the
case of SR, the fundamental mathematical building blocks from which the model
form is constructed are defined by the user.

From a general perspective, SR is an optimization problem that occurs over a
non-numeric domain of mathematical operators.  There are binary operators which
utilize two operands (\eg $+$, $-$, $\times$, $\div$) and functions (unary
operators) which utilize one operand (\eg $\sin$, $\cos$, $\exp$, $\ln$).
An SR model is then characterized by a variable-length combination of these
operators and coefficients and, therefore, poses an infinite space of possible
model forms to search.  In practice, the mathematical operator domain is limited
by a finite set of operations and maximum model complexity threshold. While
these practical constraints help constrain the vast search space an efficient
search method is required to discover accurate models.  

To search the space, genetic programming (GP) is the most commonly used approach to
SR, together termed GPSR.  Within GPSR, genetic algorithms are used to evolve
models based on their fitness relative to a specified fitness function(s), which
are discussed in the next section.  This fitness function is used to select models most
likely to perform better, after which model evolution occurs through random
recombination (\ie crossover) and permutation (\ie mutation) to generate new
candidate models. At the same time, the candidates with poorest fitness are
dropped out of the population (\eg natural selection).  The iterative
exploration of the solution space is subject to both randomization and guidance
from the particular fitness and crossover and mutation procedures implemented.

Potential GPSR fitness functions include standard explicit error metrics (\eg
mean squared error), custom reward-cost functions, or derivative-based fitness
functions for implicit equations
\cite{schmidt2010symbolic,bomarito2021development}. Combining these with
alternatives for solution representation/structure (acyclic graphs) and
evolution strategies make GPSR suitable to a variety of applications. However,
the expense for this flexibility and benefits includes increased computational
resources, potential non-determinacy, and susceptibility to selecting
high-variance solutions \cite{Wang_gpsr_2019}. 

Here, we use the open-source NASA GPSR code Bingo \cite{bingo, bingo_paper}
because of its modular nature, allowing for implementation of custom fitness
functions, and its ability to scale to high performance computing (HPC)
resources.  To improve the performance and robustness within Bingo, several
efforts are made, such as deterministic crowding \cite{mahfoud1995niching} and
parallel island evolution \cite{fernandezi20056,whitley1999island}.
Deterministic crowding is a common niching algorithm to avoid converging to
local optima wherein pairs of models are generated based on their similarities,
and best fit individuals survive to the next generation. Further, we implement
a parallel island evolution strategy whereby multiple islands (archipelago) are
distributed across available computer cores.  Individual islands then
independently evolve with periodic communication of models among the islands to
help maintain diversity.

\subsection{Physics-regularized fitness function}
\label{sec:fitness}
Setup begins with a set of training data $(X_i, y_i)$ where $i \in
{1,2,\ldots,n}$ and $X_i$ is a \emph{p}-dimensional vector-valued input of
features, $X_i={x^{(1)}_i, x^{(2)}_i, \ldots, x^{(p)}_i}$, and $y_i$ are the
corresponding labels. Hereafter, the index on $X$ is dropped for brevity as a
feature vector is always implied. A model, $f:\mathbb{R}^p \rightarrow
\mathbb{R}$, is then sought for these training data.  For each model proposed by
GPSR, $\tilde{f}(X)$, a defined fitness function quantifies its accuracy and is
sought to be minimized.  Conventionally, a purely data-driven fitness, $F^{dd}$,
would be defined as a vector: 

\begin{equation}
  F^{dd}_i = \tilde{f}(X) - y_i
  \label{eqn:F1i}
\end{equation}
or homogenized as, for example, a mean-squared error:

\begin{equation}
  F^{dd} = \frac{1}{n} \sum_{i=1}^n (F^{dd}_i)^2.
  \label{eqn:F1}
\end{equation}

To impose physics regularization, the fitness is augmented with a measure of how
well $\tilde{f}$ satisfies the prescribed differential equations,
$L^{(k)}(\tilde{f}(X^{(k)}))$, where $L$ is an arbitrary differential operator.
Here, $k \in {1,2,\ldots,l}$, represents $l$ differential equations \eg various
boundary conditions or multiple governing physics. Further, the $X^{(k)}$ at
which these differential equations are evaluated need not be
coincident with the purely data-driven inputs, $X$, nor consistent across the
$L^{(k)}$.  A purely physics-regularized fitness, $F^{pr}$, can then be defined
as:

\begin{equation}
  F^{pr}_j = [\lambda^{(1)} L^{(1)} (\tilde{f}(X^{(1)}));
               \lambda^{(2)} L^{(2)} (\tilde{f}(X^{(2)})); \ldots
               \lambda^{(l)} L^{(l)} (\tilde{f}(X^{(l)}))],
  \label{eqn:F2j}
\end{equation}
\noindent
where the semi-colon indicates concatenation into a one dimensional vector and $j \in
{1,2,\ldots,m}$, where $m$ is the total number of $X^{(k)}$ data points.  Similar
to Equation \ref{eqn:F1}, this vector-valued fitness can then be homogenized as,
for example, a mean-squared error.  This second, physics-regularized, fitness
term includes optional hyperparameters, $\lambda^{(k)}$, which control a
relative weighting among the conventional training data and the $k$ differential
equations.  It is often sufficient to set all $\lambda^{(k)} = 1$, as is done
herein, and results in a complete, concatenated, vector-valued fitness:

\begin{equation}
  F_{n+m} = [F^{dd}_i; F^{pr}_j],
  \label{eqn:F}
\end{equation}
\noindent
or, can also be homogenized as, for example, a mean-squared error which is
referred to hereafter as $F$.

To evaluate the physics-regularized fitness term, $F^{pr}$,
$L^{(k)}(\tilde{f}(X_j))$ must be computed.  As described in the previous
section, Bingo produces a set of mathematical operators that compose each
function.  To enable the computation of the requisite derivatives, we simply
translate these mathematical operators into primitives provided by 
PyTorch\footnote{Specific vendor and manufacturer names are explicitly mentioned only to accurately describe the test hardware. The use of vendor and manufacturer names does not imply an endorsement by the U.S. Government nor does it imply that the specified equipment is the best available.}
\cite{NEURIPS2019_9015} and employ the automatic differentiation method,
autograd \cite{captum2019github}.  Automatic differentiation is a mature
technology that has gained further prominence in recent years due to its use in
training neural networks (NN).  However, the NN use-case and our own differ in
an important way. NNs construct a single, highly-parameterized trial function.
The resulting gradients are then correspondingly high dimensional, resulting in
a higher computational demand.

Automatic differentiation frameworks tend to be designed for NNs and, therefore,
favor spending greater time when the function is first constructed to make
repeated computation of the gradient more efficient. In our case, however, we
construct a population of analytic functions for which the derivative will be
computed.  The dimensionality of the necessary derivatives are also much lower
(\eg 3-4 dimensions for PDEs), and the derivatives will only need to be
evaluated once per function construction. Thus, techniques such as extensive
graph-optimization and JIT (just in time) compilation tend to harm rather than
help performance in our case. PyTorch was the fastest framework which we tried;
however, it is still slower than conventional function evaluation by an order of
magnitude. It is possible that algorithmic differentiation can be sped up
significantly via using a framework tailored specifically for this use-case.

\subsection{Local optimization of coefficients}
\label{sec:clo}
Once a model has been formed via GPSR as in Section \ref{sec:gpsr}, commonly
there are undetermined constant coefficients.  The physics-regularized fitness
defined in the previous section is then sought to be minimized by subjecting
these constant coefficients to a local optimization step. The performance of
physics-regularized GPSR is found to be highly sensitive to this local
optimization step.  As part of this work, many of the optimization algorithms
made available through the scipy.optimize.minimize and scipy.optimize.root
modules in Python were tested \cite{SciPy}. For vector-valued fitness functions,
Equation \ref{eqn:F}, root-finding methods are utilized while for scalar-valued
(homogenized) fitness functions the minimization methods are utilized.  Because
this local optimization is automated for every GPSR model there is no
model-specific customization of the initial guesses.  Instead, initial guesses
for the constant coefficients were selected randomly from a uniform distribution
from -1 to 1. However, optimized parameters were unconstrained and, as such,
could venture outside those initial guess bounds.

Because GPSR evolves arbitrary functions this assessment of available methods
tests which optimization methods perform best in general.  Of the available
methods in scipy 1.8.1, the Levenberg–Marquardt (LM) algorithm performed best
for vector-valued fitness functions while the Broyden–Fletcher–Goldfarb–Shanno
(BFGS) algorithm performed best for scalar-valued fitness functions
\cite{SciPy}. LM is a common optimization method used for non-linear least
squares problems. BFGS is an iterative method used for unconstrained, nonlinear
optimization problems. As with many gradient-based optimization methods, both LM
and BFGS can converge on local minima. Beyond simply performing best, it was
found that LM and BFGS algorithms were the only two algorithms that frequently
found the global optima and ran approximately $10 \times$ faster than other
methods. This observation is consistent with previous findings
\cite{de2015evaluating}.  Ultimately, LM had approximately the same runtime as
BFGS but more frequently determined the global optimum so it was selected for
all of the numerical experiments presented next.  The improved performance of
LM is likely related to preservation of the vector-valued fitness function,
where homogenization of fitness to a scalar value is not required.

\section{Experiments}
\label{sec:experiments}
Two numerical experiments based on boundary-value problems were selected to
verify that physics-regularized GPSR can evolve known solutions to differential
equations.  The first experiment verifies the solution to a linear, fourth-order
ODE. The second experiment then verifies a linear, second-order PDE. Of
particular importance with these experiments is that success occurs specifically
when the known analytic solution to these differential equations evolves \ie an
algebraic equivalence that is more rigorous than a conventional numerical
threshold. 
\subsection{Euler-Bernoulli Differential Equation}
\label{sec:eb_expr}
\begin{figure}[h]
    \centering
    \includegraphics[width=5cm]{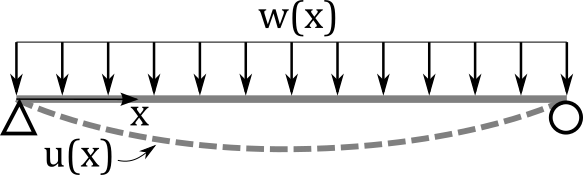}
    \caption{Simply-supported beam with uniform load.}
    \label{fig:beam_bending}
\end{figure}

The first numerical experiment is a fourth-order ODE boundary value problem,
which is derived from force and moment equilibrium, and known as the
Euler-Bernoulli equation. The objective of this experiment is to verify that
physics-regularized GPSR can reliably evolve the known analytical solution,
$u(x)$, to the known ODE:

\begin{equation}
	L^{(1)}(u(x)) = \frac{\partial^{4} u(x)}{\partial x^{4}} - \frac{w(x)}{EI} = 0,
  \label{eqn:eb_ode}
\end{equation}
\noindent
where $u$ is the deflection of the beam under applied load, $w$, along its
length, $x$, E is the Young's modulus, and I is the moment of inertia. The
boundary conditions are such that the $u(x=0)=0$ and $u(x=l)=0$.  Similarly, the
curvature, $L^{(2)}=\frac{d^2u}{dx^2}-\kappa$=0, is such that $\kappa(x=0)=0$
and $\kappa(x=l)=0$.  For this experiment, $w(x)$ is taken to be a uniform load,
Figure \ref{fig:beam_bending}, and we rewrite the constant term of Equation
\ref{eqn:eb_ode} as $\frac{w(x)}{EI}=c$.

The solution to Equation \ref{eqn:eb_ode} can be readily obtained by
integration and application of boundary conditions, upon which Equation
\ref{eqn:eb_soln} is obtained:

\begin{equation}
    u(x) = \frac{c}{24} (x^{4} - 2lx^{3} + l^{3}x).
  \label{eqn:eb_soln}
\end{equation}

Representing practical values, we define $c=\expnumber{5}{-5}$ and $l=10$, which
results in the expanded form given in Equation \ref{eqn:soln}. This equation
consequently serves as a model form with specific coefficients that will be used
to verify the produced GPSR models. Because of the symbolic nature of GPSR, we
not only verify numerical accuracy, but also algebraic equivalence between the
known solution and GPSR models. Note, this ability to verify model form and
coefficients represents the fundamental difference between physics-regularized
GPSR and its black-box methods counterparts, \eg PINNs, and is a necessary step
for engineering use cases.

\begin{equation}
	u(x) = 2.08\overline{33} \times 10^{-6} x^{4} - 4.1\overline{66} \times 10^{-5} x^{3}
          + 2.08\overline{33} \times 10^{-3} x .
    \label{eqn:soln}
\end{equation}

This Euler-Bernoulli problem was run with $n = (2, 3, 11)$ input training data
pairs $(x_i, y_i)$ to assess the relative importance of physics-regularization
to provided training data quantity. In the case of $n=2$ training points, this
represents specification of only the boundary conditions stated for $u(x)$.
This numerical experiment is extended to a single training point at the center
of the beam for $n=3$ training points and ultimately to a training point at
each unit distance along the beam with $n=11$ training points. For each of the
physics-regularized trials, $x^{(1)} \in (0,l)$ and $x^{(2)} \in
(1,2,\ldots,9)$: a total of $m=11$ physics-regularization points.  

As a baseline performance comparison for physics-regularized GPSR, conventional
GPSR was also completed.  For the conventional GPSR trials, training data sizes
of $n = (3, 5, 11)$ were tested.  Because the case of $n=2$ training points
could not be carried out with conventional GPSR (as the solution would be
trivial, $f(x)=0$) a case of $n=5$ training points was added to illustrate data quantity-dependent performance behavior. 

\begin{table}[h]
  
    \centering
    \caption{GPSR hyperparameters for solving the Euler-Bernoulli equation.}
    
    \begin{tabular}{c|c|c}

    \hline
    \multicolumn{2}{c|}{\textbf{Hyperparameters}} & \textbf{Value(s)}\\
         \hline
         \multirow{2}{*}{Operator}&Test 1& $+$, $-$, $\times$\\
         \cline{2-3}
          &Test 2& $+$, $-$, $\times$, $pow$, $\sin$, $\div$\\
         \hline
         \multicolumn{2}{c|}{Number of islands} & 10\\
         \hline
         \multicolumn{2}{c|}{Population size} & 150\\
         \hline
         \multicolumn{2}{c|}{Maximum complexity} & 10\\
         \hline
         \multicolumn{2}{c|}{Crossover rate} & 0.5\\
         \hline
         \multicolumn{2}{c|}{Mutation rate} & 0.5\\
         \hline
         \multicolumn{2}{c|}{Differential weight, $\lambda$} & 1\\
         \hline
         \multicolumn{2}{c|}{Evolutionary algorithm} & Deterministic crowding\\ 
         \hline
         
    \end{tabular}
    \label{tab:eb_hyper}
\end{table}

Because of the stochastic evolutionary process inherent in GP, each of these
cases was repeated 30 times to determine average behavior.  For each of these
numerical experiments the hyperparameters listed in Table \ref{tab:eb_hyper}
were used. These hyperparameters were not chosen to optimize the performance of
GPSR for this case.  Instead, generic defaults were used \eg 50\% crossover vs.
mutation, to demonstrate functionality in a generic sense, \ie without a bias
toward calibrated hyperparameters. Further, we test dependence of successful
solution evolution on the user-defined mathematical building blocks (operators).
In Test 1, operators are limited to the minimal requisite set from which the
solution can be evolved. In contrast, Test 2 permits an equal number of
unnecessary operators, which expands the models search space for GPSR.

\subsection{Poisson's Equation} 
\label{sec:ps_expr}
\begin{figure}[h!]
    \centering
    \includegraphics[width=0.4\linewidth]{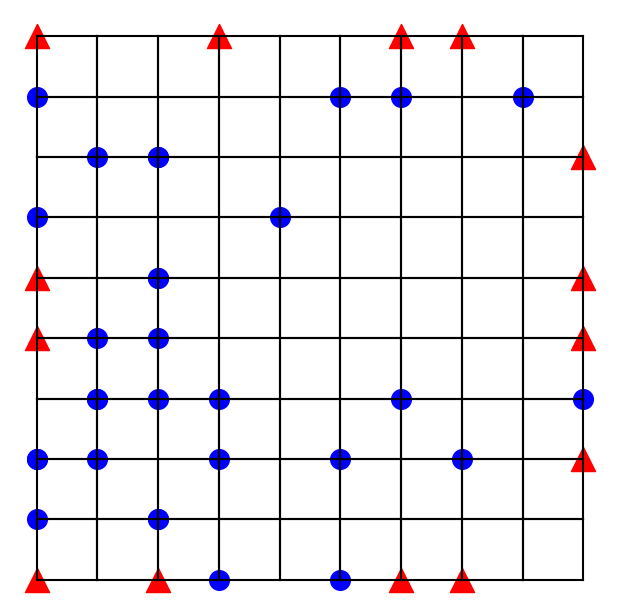}
    \caption{Example of the sampled data in the 2D domain}
    \label{fig:sampled_2d_domain}
\end{figure}

The next test problem is an elliptic PDE boundary value problem known as
Poisson's equation, which has a variety of applications in theoretical physics
\cite{evans2010partial,han2011elliptic}. The objective of this experiment is to
evolve the known closed-form solution using physics-regularized GPSR.
Additionally, we investigate the influence of domain dimensionality, \ie one,
two, or three dimensional (1D, 2D, 3D, respectively) and permitted operators
(similar to the Euler-Bernoulli experiment) on the performance of
physics-regularized GPSR. Poisson's equation with Dirichlet boundary conditions is defined as:

\begin{subequations}
  \begin{gather}
   \label{eqn:poisson_gov}
    L^{(1)}(u(x)) = \nabla^2 u(x) = f(x)  \quad\text{ in } \Omega\\
    \label{eqn:Dirichlet}
    u(x) = 0 \quad\text{ on } \partial\Omega,
  \end{gather}
  \label{eqn:poisson_pde}
\end{subequations}
where $u(x)$ is the solution on $x
\in (0,1)^d$, $f(x)$ is a source term defined by $-d~ \pi^2\prod_{i=1}^d
sin(\pi x_{i})$ where $d \in (1,2,3)$, $\Omega$ is the problem domain, and $\partial\Omega$ is the boundary of the problem. Figure \ref{fig:sampled_2d_domain} is an example of the sampled coordinates in the 2D domain, where Equation \ref{eqn:poisson_gov} is evaluated at circles and Equation \ref{eqn:Dirichlet} is evaluated at triangles. 

The solution to Equation
\ref{eqn:poisson_pde} can be obtained through integration and application of
boundary conditions, which leads to:

\begin{equation}
	u(x) = \prod_{i=1}^d sin(\pi x_{i})
  \label{eqn:poisson_solution}
\end{equation}

The analytical solutions for 1D, 2D and 3D then become Equation
\ref{eqn:poisson1D_solution}, \ref{eqn:poisson2D_solution} and
\ref{eqn:poisson3D_solution}, respectively. The resulting physics-regularized
GPSR models were evaluated numerically and symbolically to confirm if the target
solution was evolved:

\begin{subequations}
  \begin{gather}
   \label{eqn:poisson1D_solution}
    u(x_{1}) =  sin(\pi x_{1})\\
    \label{eqn:poisson2D_solution}
    u(x_{1},x_{2}) =  sin(\pi x_{1})\cdot sin(\pi x_{2})\\
    \label{eqn:poisson3D_solution}
    u(x_{1},x_{2},x_{3}) =  sin(\pi x_{1})\cdot sin(\pi x_{2})\cdot sin(\pi x_{3}) .
  \end{gather}
\end{subequations}

A numerical experiment was run with randomly sampled coordinates along the
boundary, $n = 2$ for 1D, $n = 16$ for 2D, $n = 20$ for 3D, for evaluation of
Equation \ref{eqn:Dirichlet}, and along the domain, $m = 2$ for 1D, $m = 32$ for
2D, $m = 64$ for 3D, for evaluation of Equation \ref{eqn:poisson_gov}, see Figure \ref{fig:sampled_2d_domain}. This tests the minimal amount of
input information for a well-posed problem.

\begin{table}[h!]
    \centering
    \caption{GPSR hyperparameters for solving the Poisson's equation.}
    \begin{tabular}{c|c|c}
    \hline
    \multicolumn{2}{c|}{\textbf{Hyperparameters}} & \textbf{Value(s)}\\
         \hline
         \multirow{2}{*}{Operator}&Test 1& $\times$, $sin$\\
         \cline{2-3}
          &Test 2& +, \textminus, $\times$, $\div$, $sin$, $cos$\\
         \hline
         \multicolumn{2}{c|}{Number of islands} & 10\\
         \hline
         \multicolumn{2}{c|}{Population size} & 150\\
         \hline
         \multicolumn{2}{c|}{Maximum complexity} & 20\\
         \hline
         \multicolumn{2}{c|}{Crossover rate} & 0.5\\
         \hline
         \multicolumn{2}{c|}{Mutation rate} & 0.5\\
         \hline
         \multicolumn{2}{c|}{Differential weight, $\lambda$} & 1\\
         \hline
         \multicolumn{2}{c|}{Evolutionary algorithm} & Deterministic crowding\\ 
         \hline
    \end{tabular}
    \label{tab:hyperparameters-Poisson}
\end{table}

For each of these numerical experiments the hyperparameters listed in Table
\ref{tab:hyperparameters-Poisson} were used to compare performance. As with the
Euler-Bernoulli experiment, the number of islands, population size, crossover
rate, mutation rate, differential weight, and evolutionary algorithm were not
tuned in an attempt to demonstrate a generic, default performance.  Also, tested
in this experiment is the dependence of performance on user-defined mathematical
building blocks (operators).  In Test 1, operators are limited to the minimal
requisite set from which the solution can be evolved.  Additionally, Test 2
permits a $3\times$ number of unnecessary operators.

\section{Results}
\label{sec:results}
Each of the numerical experiments presented in this section were run 30 times to
gather variation in results due to the inherent stochasticity of the genetic
programming algorithm. Each GPSR trial was terminated upon evolving a model for
which $F \le \expnumber{1}{-15}$, see Equation \ref{eqn:F}.  Of particular
interest here is the ability of GPSR to evolve the known (correct) equation, not
just a numerically-accurate approximation to that equation.  Here, a correct
equation is defined as being an algebraic equivalent to the known solution and a
trial was only considered to be successful if such an equation evolved.
Consequently, success rates pertain to the frequency at which the correct
analytical solution is determined, and not by simply reaching the prescribed
fitness threshold.

\subsection{Euler-Bernoulli equation} \label{sec:eb_results} 
\begin{figure}[h!] 
  \centering
    \includegraphics[width=.85\linewidth,height=.5\linewidth]{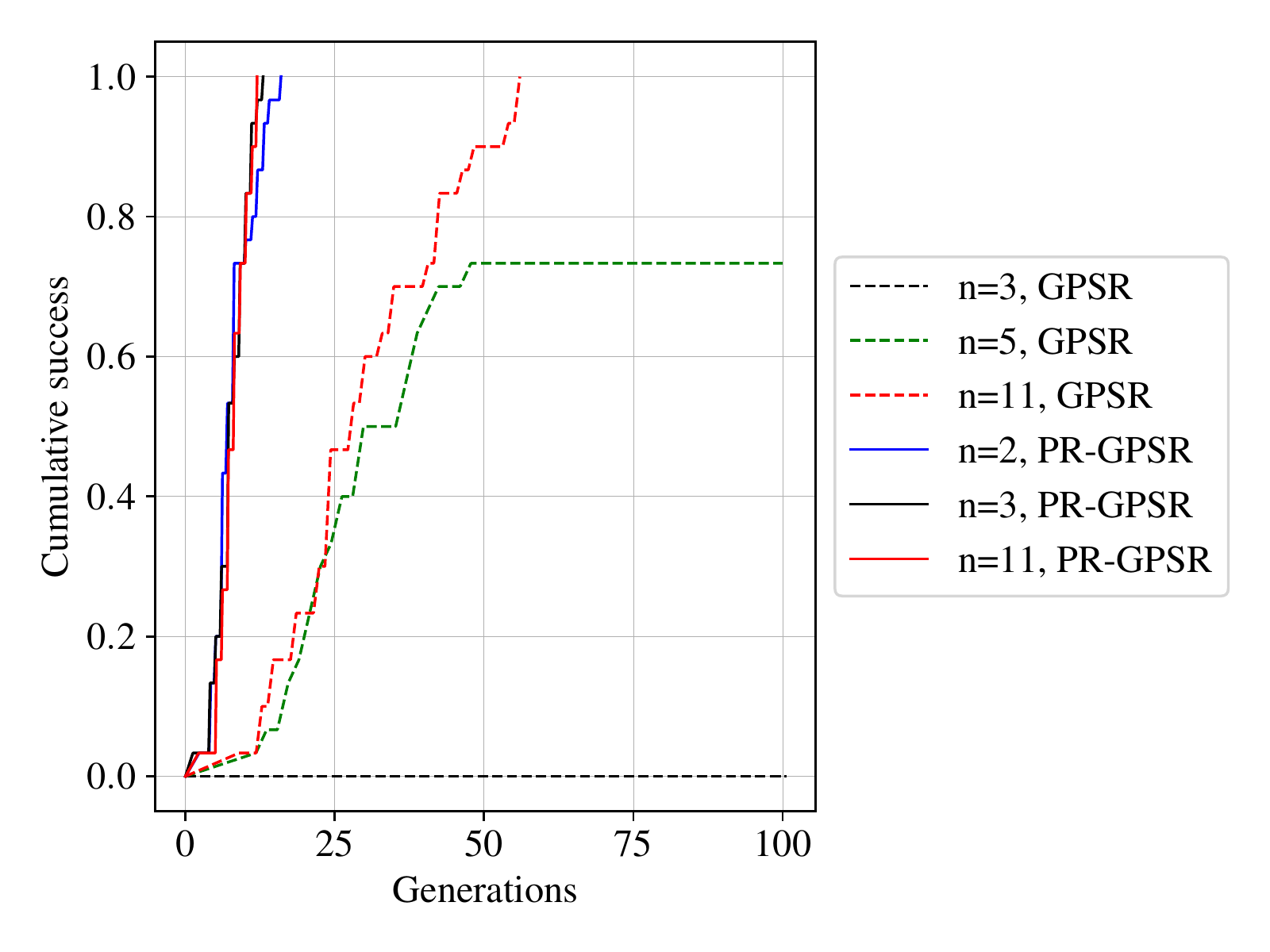} 
    \caption{Cumulative distribution of successful (\ie produced known model)
GPSR model evolution.} 
  \label{fig:all_evo} 
\end{figure}

To compare the performance of physics-regularized GPSR to conventional GPSR the
cumulative distribution of required generational counts to achieve the correct
equation is illustrated in Figure \ref{fig:all_evo}. In this experiment, the
number of points, $n$, used to evaluate the fitness contribution from Equation
\ref{eqn:F1i} is varied to quantify the effect of available training data on
successful evolution of the known solution. For the physics-regularized trials,
the number of points, $m$, used to evaluate the fitness contribution from
Equation \ref{eqn:F2j} is held fixed at 11, unless otherwise specified.  

The inclusion of physics-regularization reliably evolved the known solution is
shown in Figure \ref{fig:all_evo}, and the corresponding performance (number of
generations to success) was significantly improved.  With the conventional
fitness function GPSR was able to reliably determine the known solution if
$n=11$ training points were provided, while with $n=5$ the known solutions was
found about 75\% of the time and with $n=3$ the known solution was never
evolved.  In the cases of conventional fitness with $n=3$ or $n=5$ training
points, GPSR produced models with low fitness but with equations that only fit
those few data points but elsewhere were poor.

\begin{figure}[h!] 
    \centering
    \includegraphics[width=\linewidth,height=.5\linewidth]{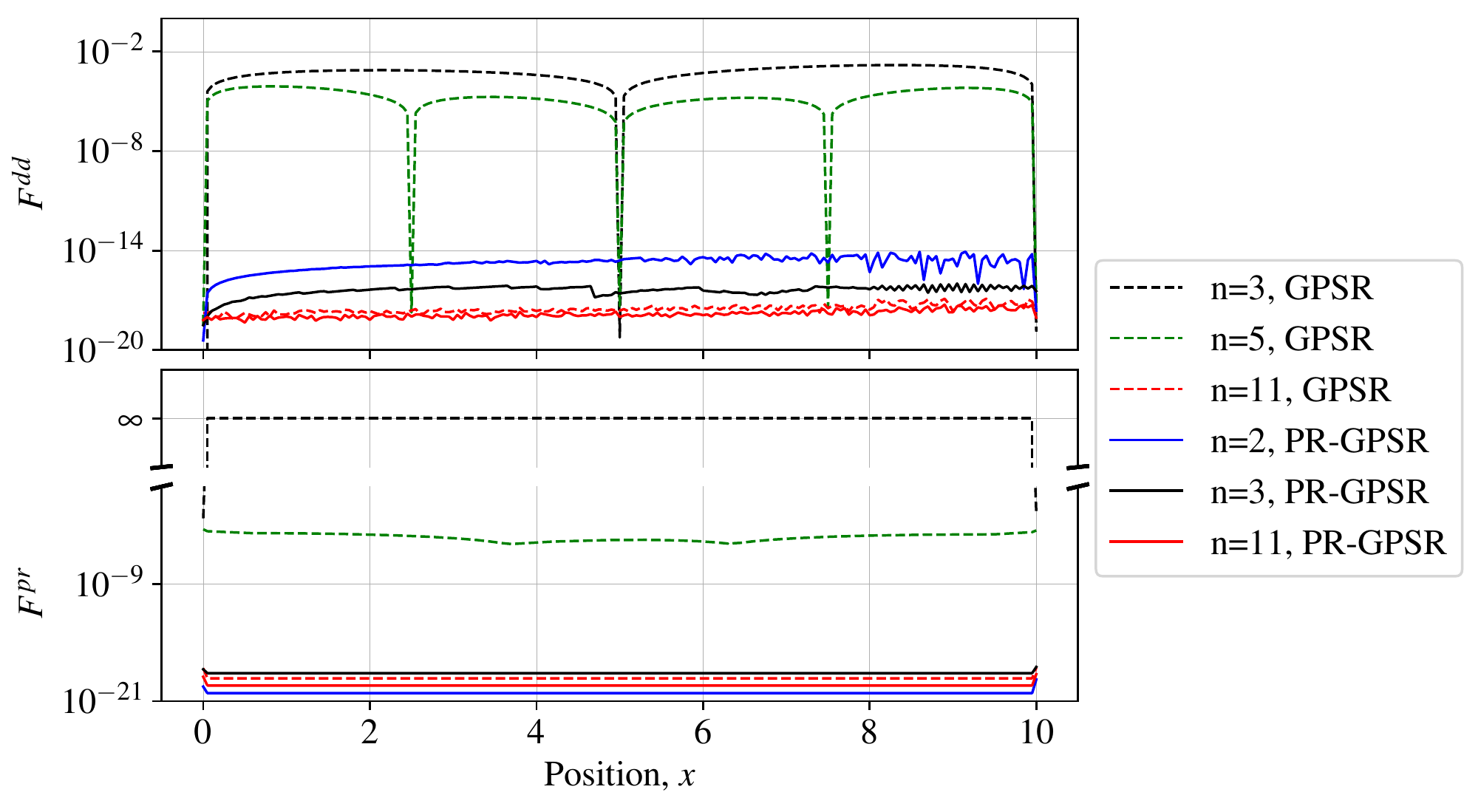}
    \caption{Test fitness evaluations at 201 points in $x$ decomposed into (top)
$F^{dd}$ and (bottom) $F^{pr}$.}
  \label{fig:pi_err} 
\end{figure}

To test and illustrate a general measure of accuracy, Figure \ref{fig:pi_err}
contains the average (of the 30 repeats) fitness terms along the beam, $x$, for
each studied case. First, the error from provided displacement training data,
$F^{dd}$, are plotted at 201 points (steps of 0.05) along the beam in Figure
\ref{fig:pi_err} (top).  As can be seen for the $n=3$ and $n=5$ training points
cases with conventional GPSR, the produced models are only accurate specifically
at the training points.  Elsewhere, those two cases produce models that are
especially inaccurate.  All other tested cases, however, produce
numerically-accurate models: producing highly accurate models at points that
were not specifically provided as training data.

Next, a test for general model
accuracy for the physics-regularized term, $F^{pr}$, is presented in Figure \ref{fig:pi_err} (bottom). For models that resulted
in undefined $F^{pr}$ evaluation \eg were not fourth-order differentiable, the
fitness was set to $\infty$.  These results are consistent with results of Figure
\ref{fig:all_evo}, where the combination of conventional fitness and low data
quantity (\ie $n=3$ or $n=5$ data points) did not reliably produce the known solution.
All other cases resulted in negligible test errors for both $F^{dd}$ and
$F^{pr}$.  For these cases, an algebraically-equivalent form of the known
solution was evolved and the non-zero error is a direct consequence of the
finite error threshold that results from the local optimization step,
Section \ref{sec:clo}, and round-off error during model evaluation.  

In both tests for general model accuracy, $F^{dd}$ and $F^{pr}$, the
physics-regularized cases worked well and performance was relatively insensitive
to the quantity of training data provided. In other words, even in the case that
only the boundary conditions (\ie $n=2$ training points) are provided,
physics-regularized GPSR produces the known solution. By contrast, with the
conventional fitness, GPSR only reliably produced the known solutions with
$n=11$ training points. Further, from these results it is observed that the
incorporation of physics-regularized fitness is even more beneficial than added
training data.  Finally, beyond these numerical evaluations, symbolic regression
affords the unique opportunity to evaluate the model form, algebraically.  These
evaluations are provided in Section \ref{sec:discussion} along with the
implications that these results suggest for engineering use of
physics-regularized, interpretable ML methods.

As provided in Table \ref{tab:eb_hyper}, the preceding tests were run with $+$,
$-$, and $\times$ as the permitted operators for GPSR.  And, while that
represents the minimal set of operators for this experiment it is important to
understand how unneeded operators might affect GPSR performance.  Consequently,
the same numerical experiment was repeated (again 30 times), but with
more operators (annotated as MO): $pow$, $sin$, and $\div$. 

\begin{figure}[h!] 
  \centering
    \includegraphics[width=0.85\linewidth,height=.5\linewidth]{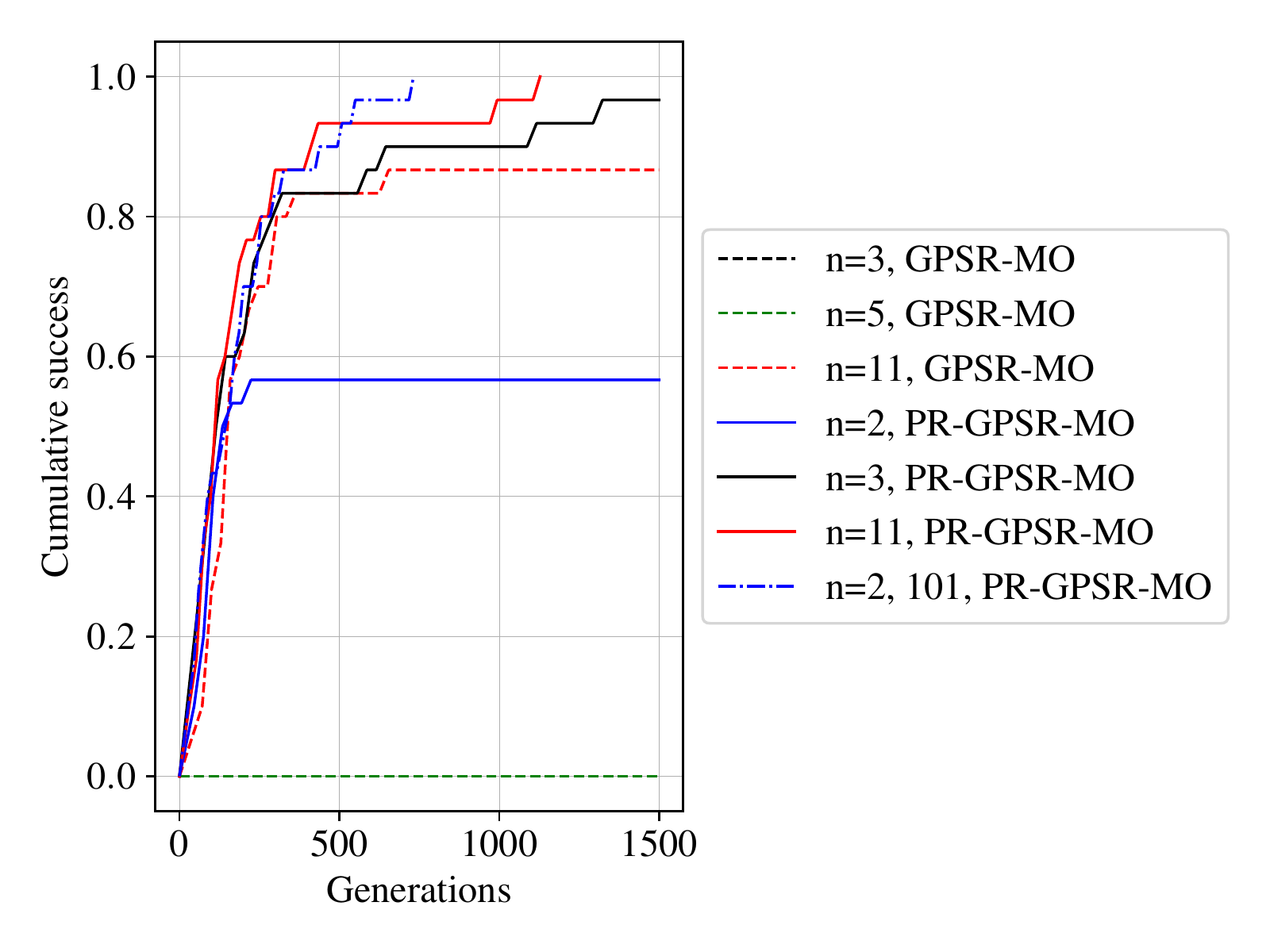} 
    \caption{Cumulative distribution of successful GPSR model evolution with
unneeded operators permitted.} 
  \label{fig:all_evo_mo} 
\end{figure}

For these tests with unneeded operators a corresponding shift in behavior is presented in Figure \ref{fig:all_evo_mo}. First, it is seen that the cases of
conventional fitness with $n=3$ or $n=5$ training points were never observed to
produce the known solutions while with 11 training points the correct solution
was evolved only approximately 85\% of the time and required approximately
$20\times$ the number of generations as compared Figure \ref{fig:all_evo}.
Next, unlike with the previous case with a minimum operator set, the
physics-regularized cases now demonstrate a clear dependence of success on
training data quantity. Here, the case of two training data points (BCs only)
now produces the known solution just over half the time and approximately 95\%
with three points.  Extending to $n=11$ training points reliably produces the
known solution, but requires approximately 100x the generations. An additional
case was assessed here, with $n=101$ training points, which demonstrates continued
performance improvement in the requisite generation count for determination of
the known solution, but still requires an order of magnitude more generations
than illustrated in Figure \ref{fig:all_evo}. This qualitative shift of
performance in reproducing the known solution was determined to mainly be a
result of including the $sin$ operator.  For evolved models that included $sin$,
$F^{pr}$ was defined (\ie always fourth-order differentiable) and the local
optimization was consistently successful in finding model parameters that
produced a low-fitness model.  However, while these models were numerically
accurate, they were not algebraically equivalent to the known solution and
therefore did not constitute success in these tests.  Clearly, the included
operators play a significant role in GPSR performance even with the inclusion of
a physics-regularized fitness function.



\subsection{The Poisson's Equation}
\begin{figure}[h!]
    \centering
    \begin{subfigure}[b]{0.23\textwidth}
        \centering
         \includegraphics[width=\textwidth,height=0.65\textwidth]{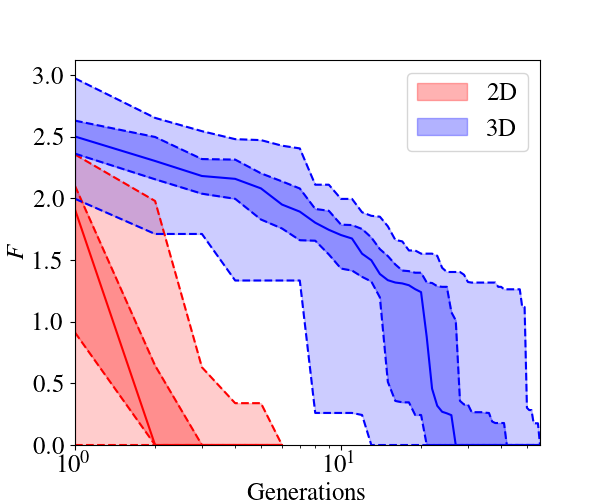}
         \caption{2D, Test 1 \& 3D, Test 1}
         \label{fig:2D-test1,3D-test1}
    \end{subfigure}
    \hfill
    \begin{subfigure}[b]{0.23\textwidth}
        \centering
         \includegraphics[width=\textwidth,height=0.65\textwidth]{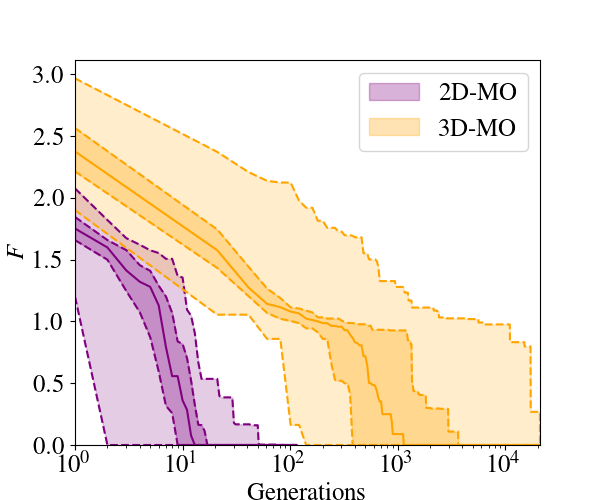}
         \caption{2D, Test 2 \& 3D, Test 2}
         \label{fig:2D-test2,3D-test2}
    \end{subfigure}
    \hfill
    \begin{subfigure}[b]{0.23\textwidth}
        \centering
         \includegraphics[width=\textwidth,height=0.65\textwidth]{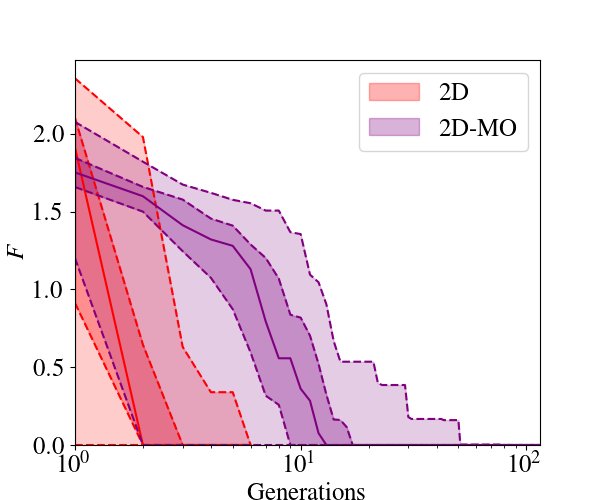}
         \caption{2D, Test 1 \& 2D, Test 2}
         \label{fig:2D-test1,2D-test2}
    \end{subfigure}
    \hfill
    \begin{subfigure}[b]{0.23\textwidth}
        \centering
         \includegraphics[width=\textwidth,height=0.65\textwidth]{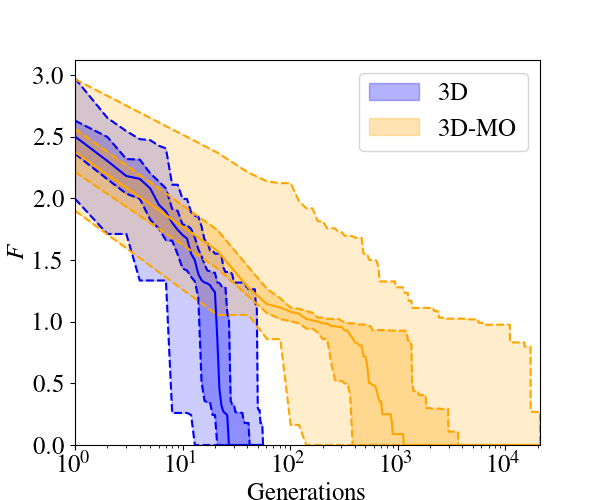}
         \caption{3D, Test 1 \& 3D, Test 2}
         \label{fig:3D-test1,3D-test2}
    \end{subfigure}
    
    \caption{Fitness results of 30 SR runs for all trials. The solid
line is the median fitness at each generation. The two filled regions of each
color represents the quartile range (darker) and outliers (lighter), respectively.}

    \label{fig:All_trend}
\end{figure}

A plot of 30 GPSR trials is shown in Figure \ref{fig:All_trend} to visualize the
influence of domain dimensionality and permitted operators on the performance of
the physics-regularized GPSR implementation.  For consistency, each result is
represented as the same color in all sub plots. The solid lines correspond to
median values and the darker and lighter colored section correspond to the
quartile range and outliers, respectively. From Figure \ref{fig:All_trend},
it is observed that all cases achieved a fitness below the specified threshold,
$F \le \expnumber{1}{-15}$. Further, in each of the trials, the known solution was
successfully evolved.  And generally, as would be expected, the requisite number
of generations increased significantly with problem dimensionality.

The effect of domain dimensionality on
the performance with only two operators ($sin$ and $\times$) is presented in Figure \ref{fig:2D-test1,3D-test1}. It can be seen
that 1 generation was required for 1D (not illustrated and not surprising
due to the low complexity of the solution), 2 generations for 2D, and 28
generations for 3D. The effect of the unneeded operators, $+$, $-$, $\div$ and $cos$, is illustrated in Figure \ref{fig:2D-test2,3D-test2}. Overall, the trend is
similar to Figure \ref{fig:2D-test1,3D-test1}, but requires significantly more
generations for success. This comparison is made more clear by Figures
\ref{fig:2D-test1,2D-test2} and \ref{fig:3D-test1,3D-test2}.  In each of these
tests, lower dimensions tended to result in lower fitness.  This result was
found to be due to fewer parameters to be determined during local optimization
and round-off error.

\begin{figure}[h!]
    \centering
    \includegraphics[width=0.85\linewidth,height=0.45\linewidth]{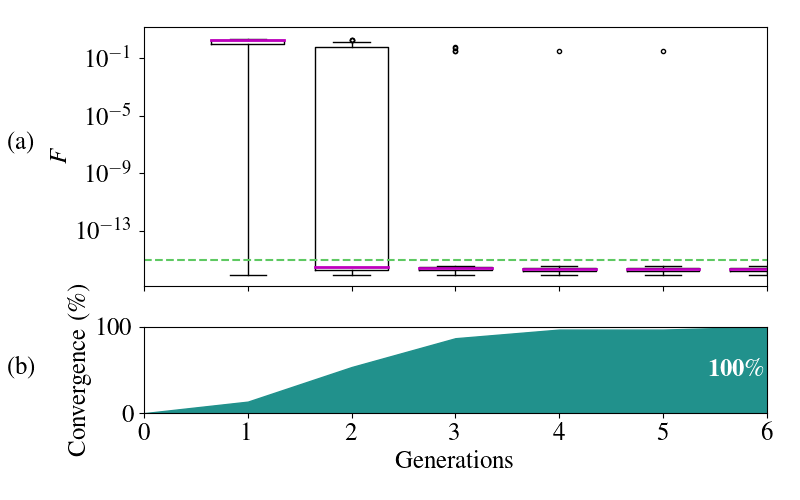}
    \caption{Successful evolution of known solution over generations in the 2D,
Test 1 trial. (a) fitness evolution as box plot of most fit individuals in each
population across the 30 runs. The median is illustrated by the purple line, and
outliers by circles. The numerical convergence threshold is shown as the green
dashed line. (b) cumulative distribution of success.}
    \label{fig:2D_simple_box}
\end{figure}
\begin{figure}[h!]
    \centering
    \includegraphics[width=0.85\linewidth,height=0.45\linewidth]{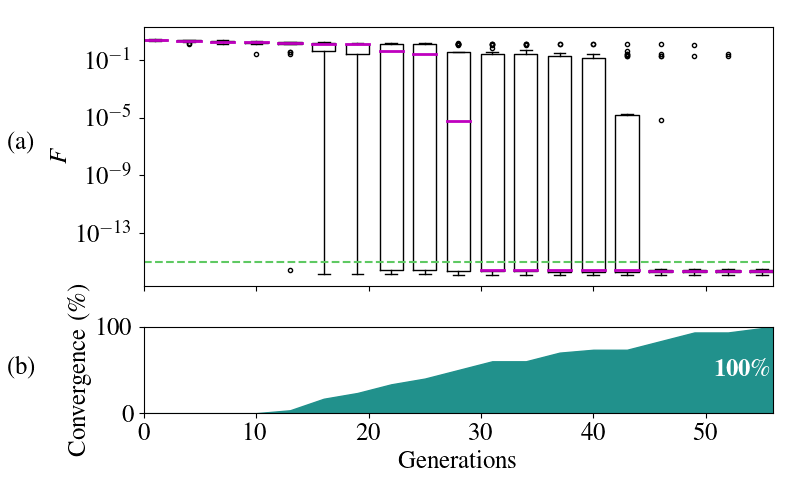}
    \caption{Successful evolution of known solution over generations in the 3D,
Test 1 trial. (a) fitness evolution as box plot of most fit individuals in each
population across the 30 runs. The median is illustrated by the purple line, and
outliers by circles. The numerical convergence threshold is shown as the green
dashed line. (b) cumulative distribution of success.}
    \label{fig:3D_simple_box}
\end{figure}
\begin{figure}[h!]
    \centering
    \includegraphics[width=0.85\linewidth,height=0.45\linewidth]{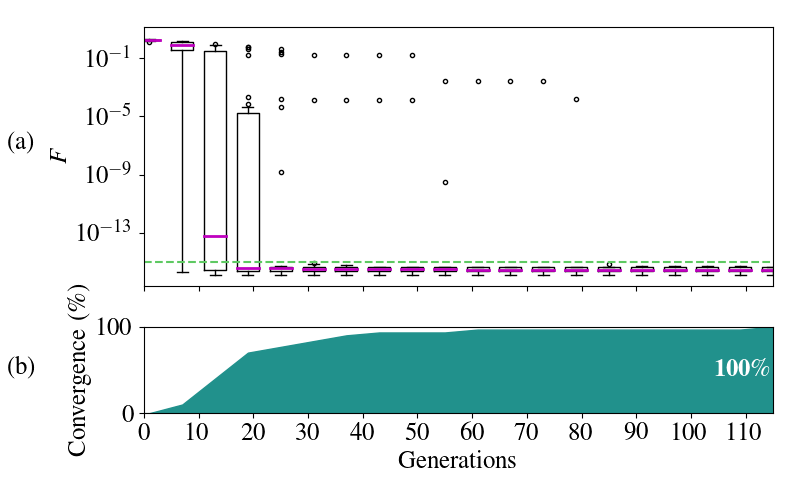}
    \caption{Successful evolution of known solution over generations in the 2D,
Test 2 trial. (a) fitness evolution as box plot of most fit individuals in each
population across the 30 runs. The median is illustrated by the purple line, and
outliers by circles. The numerical convergence threshold is shown as the green
dashed line. (b) cumulative distribution of success.}
    \label{fig:2D_all_box}
\end{figure}
\begin{figure}[h!]
    \centering
    \includegraphics[width=0.85\linewidth,height=0.45\linewidth]{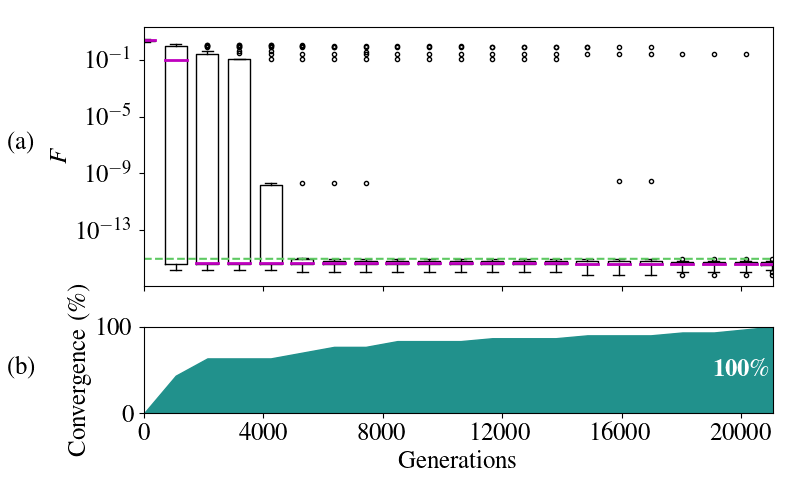}
    \caption{Successful evolution of known solution over generations in the 3D,
Test 2 trial. (a) fitness evolution as box plot of most fit individuals in each
population across the 30 runs. The median is illustrated by the purple line, and
outliers by circles. The numerical convergence threshold is shown as the green
dashed line. (b) cumulative distribution of success.}
    \label{fig:3D_all_box}
\end{figure}

The statistical effect of dimensionality and included operators on success rates is illustrated in Figures \ref{fig:2D_simple_box}-\ref{fig:3D_all_box}.  In each,
the box plot illustrates the distribution of best fit individuals in the
population across all 30 runs of each trial (combined 2D or 3D, and operator
set).  The boxes extend from the lower to upper quartile with whiskers extending
to show the distribution range, and outliers shown as open circles. For the 2D
trial with only necessary operators included (Test 1), Figure
\ref{fig:2D_simple_box}, it is seen that convergence occurs quickly with all 30
runs producing the known solution by the sixth generation. With the same set
of permitted operators (Test 1), but extended to 3D, a significant $(\sim10\times)$ increase in the number
of required generations to reliably evolve the known solution is illustrated in Figure
\ref{fig:3D_simple_box}. Next, the effect of an
increased set of permitted operators (Test 2) is presented in Figures \ref{fig:2D_all_box} and \ref{fig:3D_all_box}.  For the 2D case with increased
operators, approximately 100
generations were required for successful evolution of all 30 runs as presented in Figure \ref{fig:2D_all_box}. Lastly,
the most complex case of 3D with increased operators, Figure
\ref{fig:3D_all_box}, required approximately 20k generations. It should be noted
that these generation counts represent the requirement to achieve success across
all 30 runs. However, in each tested case, 1-2 outliers caused a
significant increase in the number of generations required for 100\% success. 

It was anticipated that higher dimensionality would increase requisite
generations for success, as illustrated.  However, data illustrated in Figures
\ref{fig:2D_simple_box}-\ref{fig:3D_all_box} highlights that 
the overall performance was more sensitive to
the inclusion of additional, unneeded operators in GPSR than the increase in
problem dimensionality.  To assess the reason underpinning this observation
consider that each GPSR model consists of \emph{d}, \emph{m}, and \emph{n}, where \emph{d} is the dimension, \emph{m} is the number of operators, and \emph{n} is the maximum complexity, respectively. An estimate of the total number of potential models in a search space is given by $\mathcal{H}$:
\begin{equation}
\begin{split}
    \mathcal{H} &= \left(2d +m\right)^{n-2d} \prod_{i=1}^{2d}\left(n - i + 1\right)\left(2d - i +1\right)\\ 
                  &\approx \mathcal{O}(\left(2d +m\right)^{n-2d}\left(2nd\right)^{2d}).
    \label{equation:cases}
\end{split}
\end{equation}
Consequently, the size of hypothesis space can be estimated for the 2D cases as
$\mathcal{O}(1.2\times10^{20})$ with minimal operators and
$\mathcal{O}(4.1\times 10^{23})$ for additional operators. Similarly, for the 3D
cases, the space is estimated as $\mathcal{O}(1.3\times 10^{25})$ for minimal
operators, and $\mathcal{O}(3.8\times 10^{27})$ for added operators. 
Overall, this estimates the relative influence of the increased dimensionality
and operator set on the GPSR search space and, as a consequence, ability to find the correct solution.

\section{Discussion}
\label{sec:discussion}
As was the goal of including the physics-regularized fitness function in GPSR,
the preceding results demonstrate meeting the more stringent ability for mathematical
verification \ie determination of known analytical models. This demonstration
is especially important in the engineering and science contexts due to the
resulting explainability of produced models and the trust that can be built.  In
this final section, we discuss various aspects of the results specifically
related to observation of evolved equation forms for the GPSR experiments.

\subsection{Euler-Bernoulli equation}
As can be inferred from Equation \ref{eqn:soln} the simplest, correct GPSR
model would be of the form:

\begin{equation}
\alpha \cdot x^4 + \beta \cdot x^3 + \gamma \cdot x,
\label{eq:gen_soln_form}
\end{equation}
where the following represent the global minimum to be determined during local
calibration:
$\alpha = 2.08\bar{3} \times 10^{-6}$,
$\beta  = -4.1\bar{6} \times 10^{-5}$, and
$\gamma = 2.08\bar{3} \times 10^{-3}$.

However, there are an infinite number of possibilities in which equivalent model
forms evolve within GPSR. Assessment of equivalency with the correct model then
requires additional efforts in automating GPSR model simplification. For this,
the python sympy module was utilized \cite{SymPy}.  In all
of the Euler-Bernoulli trials we observed that an algebraically-equivalent model
form, to Equation \ref{eq:gen_soln_form}, evolved in cases where the general
fitness test, see Figure \ref{fig:pi_err}, was $< \expnumber{1}{-15}$ across
the 201 test points.  Note that this threshold was found to be conservative but
ensured no false positives in the verification tests. Additionally, in the case
of the minimal set of operators, physics-regularized GPSR produced the correct
model form and coefficients reliably while conventional GPSR produced the
correct model only if sufficient training data were provided for
determination of the necessary 4$^{th}$-order polynomial. One such observed
example is:

\begin{equation}
(b\cdot d\cdot f)\cdot x^4 + (b\cdot d\cdot e)\cdot x^3 + (b\cdot c)\cdot x + a.
\end{equation}
In which case, local optimization can successfully return:

\begin{align}
\label{eq:success_clo}
b\cdot d\cdot f &= \alpha = 2.08\bar{3} \times 10^{-6}, \nonumber \\
b\cdot d\cdot e &= \beta = -4.1\bar{6} \times 10^{-5}, \\
b\cdot c &= \gamma = 2.08\bar{3} \times 10^{-3}, \text{ and}  \nonumber\\
a&=0. \nonumber 
\end{align}

Upon analyzing the evolved models that had not yet achieved the specified fitness
threshold, it was commonly observed that the correct model form had evolved but
with more complex interrelationships among the coefficients.  In these cases,
successful local optimization as given in Equations \ref{eq:success_clo} was
precluded. For example, the GPSR-produced model:

\begin{equation}
(b^2\cdot d)\cdot x^4 + (2\cdot a \cdot b \cdot d)\cdot x^3
+ (a^2\cdot d)\cdot x^2 + (a\cdot c)\cdot x
\label{eq:compl_sol_form}
\end{equation}
would have the correct model form if the $x^2$ coefficient, $a^2\cdot d$,
evaluates to zero.  However, for this to be the case either $a$ or $d$ must be
zero, which would mean that the coefficient of $x^3$ must also become zero, ultimately leading
to an incorrect model.  For improved use in engineering and science applications, an evolutionary mechanism that algebraically simplifies the model and aggregates
these redundant coefficient relationships could improve performance and improve
interpretability, by reducing unnecessary complexity of the evolved models.
However, such simplifications would also alter evolutionary paths.

For the physics-regularized GPSR tests with two training points and unneeded
operators included, incorrect model forms like the following were often
observed:

\begin{equation}
a + b\cdot sin(c + sin(x/c - d + e/c)).
\label{eq:compl_sol_form}
\end{equation}
In these specific cases, sinusoidal models commonly resulted in low fitness due
to the even spacing of the training points, $X_i$, and differential equation
evaluation points, $X_{j}$.  It was observed that increases in either $X_i$ or
$X_{j}$, along with selecting points randomly within the problem domain
precluded these misleading models during evolution. Consequently, a convergence
test is useful in determining a true model form, \ie assessment of the
sensitivity of model form to added $X_i$ or $X_{j}$ points. Upon convergence in
this context, model forms among repeated GPSR trials should reliably contain
consistent operators.

\subsection{Poisson's Equation}
The efficacy of physics-regularized GPSR for evolving analytic solutions of PDEs
is investigated using the Poisson's equation. In this experiment, focus is on the
influence of domain dimensionality and permitted operators on successful model
evolution.  First, it is found that the physics-regularized GPSR implementation
successfully obtained the correct symbolic equation for all problem dimensions
tested, such as Equation \ref{eqn:poisson1D_solution},
\ref{eqn:poisson2D_solution}, and \ref{eqn:poisson3D_solution}. However, as is
also observed in the Euler-Bernoulli experiment, a variety of equivalent model
forms evolved, which are commonly of higher complexity than the simple form:

\begin{align}
\label{eq:poi_gen_soln_form}
\tilde{u}(x) = \prod_{i=1}^d a_i\cdot \sin\left(b_i x_i\right),\\
a_i = 1 \text{ , } b_i = \pi \nonumber.
\end{align}

Further, as the dimensionality increased, higher complexity versions of the true
model became more commonly observed. Of these cases, forms that can be
equivalent to the known solution, upon successful local optimization, were
commonly observed:

\begin{align}
\label{eq:poi_compl_1}
\tilde{u}(x) &=\prod_{i=1}^d a_i\cdot \sin\left(b_i x_i + c_i\right), \text{ where}\\
\label{eq:poi_compl_2}
a_i &= 1 \text{, } b_i = \pi \text{, } c_i \approx 0 \nonumber \text{ or } \\
\tilde{u}(x) &=\prod_{i=1}^d a_i\cdot \cos\left(b_i x_i + c_i\right), \text{ where}\\
a_i &= 1 \text{, } b_i = \pi \text{ , } c_i = -\pi/2. \nonumber
\end{align}

Unlike the Euler-Bernoulli experiment, however, this experiment always resulted
in model forms \eg Equations \ref{eq:poi_compl_1} and \ref{eq:poi_compl_2},
for which the parameters could be correctly determined upon local optimization.
Generally, upon assessment of the various model forms in both experiments, it is
evident that model bloat was increased in the polynomial solutions of the
Euler-Bernoulli experiment.

\section{Conclusions} \label{sec:conclusions} 

The developed combination of an interpretable ML method, GPSR, with fitness
regularized by known governing differential equations, is demonstrated to
discover their analytical solutions.  This method is demonstrated on two
governing differential equations, a fourth-order ODE and second-order PDE, where
the known analytical solution is successfully discovered in both cases.
Because of the inherent interpretabilty, determination of success is defined as
learning the true algebraic solution, and not limited to a numerical
approximation, as is more conventional ML methods.

The physics-regularized GPSR method requires only a statement of the (known)
governing differential equation and boundary conditions sufficient for a
well-posed problem, without need for additional training data to successfully
determine the solutions.  However, the success of the physics-regularized GPSR
method is demonstrated to be dependent on several factors.  In the
Euler-Bernoulli problem (fourth-order ODE), the analytical solution was
determined in each of the 30 repeated trials when only the requisite
mathematical operators were permitted.  However, upon expanding to twice the
number of operators to include unnecessary operators, it is found that a minimal
amount of training data points were required to guide the physics-regularized
GPSR to the correct model form.  In other words, in the practical case where the
operators in the solution to the differential equation are unknown it is still
likely that some amount of (albeit a significantly reduced quantity) training
data will be needed to identify the true model form.  In the Poisson problem
(second-order PDE), the effect of dimensionality was tested. As expected for
this case, the required number of generations to discover the true solution was
significantly increased with dimension.  Nevertheless, the analytical solution
was recovered in every trial.

The ability to use interpretable ML to discover analytical solutions to known
differential equations has broad application within computational mechanics.
Determination of an analytical expression, as opposed to
numerically-approximated data, enables broader mathematical treatments \eg
sensitivity studies, optimization, and deeper insights into the solution
characteristics, to name a few.  While the method presented here provides the
first necessary step in applied mechanics and engineering, verification, the
next step is to employ this method for cases with known differential equations
but unknown solutions.  A key to building trust within the engineering community
for using ML in practice will then be the demonstration of how one can validate
those solutions.

\section*{Acknowledgements}
The support and resources from the Center for High
Performance Computing at the University of Utah are gratefully acknowledged.
This research was sponsored in part by the Army Research Laboratory (ARL) under
Cooperative Agreement Number W911NF-12-2-0023 and by Sandia National
Laboratories under Agreement 2262518.  The second and fourth authors were
partially supported by the Defense Advanced Research Projects Agency (DARPA)
through the Transformative Design (TRADES) program under the award
HR0011-17-2-0016.  The views and conclusions contained in this document are
those of the authors and should not be interpreted as representing the official
policies, either expressed or implied, of ARL or the US Government.  The US
Government is authorized to reproduce and distribute reprints for Government
purposes notwithstanding any copyright notation herein.





\section*{Bibliography}

\bibliographystyle{elsarticle-num-names} 
\bibliography{references}

\end{document}